# Beyond designer's knowledge: Generating materials design hypotheses via large language models

Quanliang Liu[1], Maciej P Polak[1], So Yeon Kim[2], MD Al Amin Shuvo[1], Hrishikesh Shridhar Deodhar[1], Jeongsoo Han[1], Dane Morgan[1], Hyunseok Oh[1]

[1]University of Wisconsin - Madison, Madison, WI, USA, 53706
[2]Massachusetts Institute of Technology, Cambridge, MA, USA, 02139

* Corresponding author: Hyunseok Oh (hyunseok.oh@wisc.edu)

**Abstract:**
Materials design often relies on human-generated hypotheses, a process inherently limited by cognitive constraints such as knowledge gaps and limited ability to integrate and extract knowledge implications, particularly when multidisciplinary expertise is required. This work demonstrates that large language models (LLMs), coupled with prompt engineering, can effectively generate non-trivial materials hypotheses by integrating scientific principles from diverse sources without explicit design guidance by human experts. These include design ideas for high-entropy alloys with superior cryogenic properties and halide solid electrolytes with enhanced ionic conductivity and formability. These design ideas have been experimentally validated in high-impact publications in 2023 not available in the LLM training data, demonstrating the LLM's ability to generate highly valuable and realizable innovative ideas not established in the literature. Our approach primarily leverages materials system charts encoding processing-structure-property relationships, enabling more effective data integration by condensing key information from numerous papers, and evaluation and categorization of numerous hypotheses for human cognition, both through the LLM. This LLM-driven approach opens the door to new avenues of artificial intelligence-driven materials discovery by accelerating design, democratizing innovation, and expanding capabilities beyond the designer's direct knowledge.

# 1. Introduction

Materials design has traditionally evolved through a blend of inductive insights from empirical knowledge and deductive predictions based on scientific principles[1–4] (**Figure S1, Supporting Information**). The generation of design hypotheses, the initial step in materials design, mostly relies on the researchers' knowledge, which is followed by validation and reinforcement of the concepts through literature survey, experimentation, or computation.[5,6] However, the increasing complexity of materials, driven by demands for improved performance and multiple functionalities, has made the hypothesis generation step increasingly challenging. The dynamic interactions among processing, structure, and properties (P-S-P) often require the simultaneous management of large amounts of information[3]. While the expanding volume of information across different materials domains provides opportunities for generating diverse and disruptive discoveries by transferring knowledge across fields—as seen in nacre-like hybrid composites[7] and high entropy catalysts[8]—it also overwhelms researchers trying to keep up with new developments.

Recently, large language models (LLMs) have demonstrated extraordinary text retrieval, generation, and reasoning capabilities in complex language tasks.[9–12] Consequently, LLMs are being actively utilized in materials science, with applications ranging from extraction of numerical[13–16] and textual[17–20] data from literature to utilizing the information for planning and guiding autonomous experiments.[21,22] Particularly, LLMs have a remarkable ability to generate novel sentences from a wide vocabulary in response to input prompts, even if they have not been specifically trained on those exact combinations.[23] This "in-context learning" capability enables the models to perform reasoning (inference), and studies are actively being conducted to explore hypothesis generation from this capability in multiple fields.[24,25] Recent efforts have demonstrated that LLMs can generate materials science hypotheses with close human guidance through chain-of-thought prompting[24] or providing related knowledge graphs or papers.[19]

In this research, we aim to leverage LLMs for generating materials design hypotheses even without explicit guidance from human experts. This capability would enable the massive integration and utilization of knowledge in hypothesis generation, far beyond what individual researchers might possess, potentially greatly improving and accelerating materials design workflows. Among various types of hypotheses and materials design challenges, we focus on a specific yet broadly important type where the non-trivial role of LLMs can be clearly assessed.

We leverage an LLM, gpt-4-1106-preview snapshot model of GPT-4, for extracting and synergistically synthesizing meaningfully distinct mechanisms between P-S-P from tens of different papers within the interested domains and external or broader domains. The LLM then creates novel interdependencies between mechanisms that are not explicitly found in the input literature, generating synergistic hypotheses that extend beyond the current knowledge of the domain of interest.

We explicitly avoid seeking hypotheses that impact properties through the simple addition of independent mechanisms. For example, we seek a hypothesis like *create more precipitates to modulate martensitic transformation, enhancing not only precipitation hardening but also transformation-induced plasticity*, rather than *create more precipitates to enhance hardening and create more martensite to enhance plasticity*. While the latter can be useful, it represents a trivial addition of known effects, barely producing new value beyond what each component contributes on its own. Conversely, interdependent effects, as in the former case, are generally not obvious and require deep domain knowledge to develop, thereby often remaining unrealized until a thoughtful expert proposes them, as the example was actually published recently in a high-impact journal.[26] Rigorous definitions of synergy require quantitative analysis to demonstrate mutual enhancement between mechanisms. However, for initial hypothesis generation, a qualitative approach focusing on the presence of a positive unidirectional interaction between two or more mechanisms can be sufficient. Therefore, in the context of materials design, we define a synergistic hypothesis as one where at least one mechanism positively influences another, without significant negative impact.

Examples discussed here include hypotheses for high-entropy alloys (HEAs) with superior cryogenic mechanical properties and halide solid electrolytes (SEs) with high ion conductivity and superior formability/malleability. Some of the LLM-generated hypotheses have been experimentally validated in recent publications in high-impact journals,[27,28] which are not available before the knowledge cutoff date of the model used in this study. In the following sections, we introduce the approach and discuss the results.

## 2. Approach

To achieve advanced hypotheses generation without expert-specific guidance, we break the process into several steps, from paper collection to system chart visualization (**Figure 1a**). **Figure**

**1b-e** illustrate the details of each step using the example of cryogenic HEA hypothesis generation. The process begins with a design request, which can be as general as a funding opportunity announcement. To address the request, **Step I** involves collecting multiple sets of papers using non-specific keywords. In this work, two sets of papers, a total of 24 papers, are collected from high-impact journals using the Web-of-Science platform with the keywords: "cryogenic, high entropy alloy" and "high entropy alloy" (**Figure 1b**). These keywords are selected based on the rationale that a designer would likely consult other HEA papers beyond the cryogenic HEA domain to find relevant ideas. The general keywords imply that the activity can be conducted by researchers unfamiliar with the specific topics or even by LLMs. The full search criteria and list of papers are in **Note S1**. Only articles available online before the gpt-4-1106-preview knowledge cutoff date (April 2023) are retrieved.

In **Step II** (**Figure 1c**), key information from the papers is extracted and organized into a structured table to be used for the subsequent steps, reducing the token count for each paper by a factor of ten; the detailed process is shown in **Figure S2**, Supporting Information. This allows the LLM to incorporate a greater volume of literature across multiple objectives, thereby expanding the design space and increasing the potential for disruptive discoveries. The table adapts the concept of a materials system chart,[2,29] a structural representation that systematically summarizes the entities in each P-S-P domain and the mechanisms between these entities in different domains,[3–5,30–33] in a linearized P-Mechanism-S-Mechanism-P format. In **Step III** (**Figure 1d**), the LLM generates hypotheses by synergistically combining mechanisms from the compiled chart, without directly incorporating the original papers. The prompt can be summarized as: "select mechanisms in different paper sets and synergistically combine them by suggesting innovative interaction" (**Figure S3, Supporting Information**). In **Step IV** (**Figure 1e**), the generated hypotheses are evaluated and categorized by the LLM. This step narrows down thousands of hypotheses to a few tens of high-quality ones, allowing human researchers to focus on promising ideas. While various criteria can be used, we evaluate hypotheses based on the level of scientific grounding to minimize speculation and the presence synergistic mechanisms (**Figure S4, Supporting Information**). Additional steps, such as system chart visualization and detailed explanation, while non-essential for hypothesis generation, are included to facilitate interaction with human researchers during subsequent design activities. The gpt-4-1106-preview model is used for **Steps II, III, and V**, and gemini-1.5-pro is utilized for **Step IV**. Detailed instructions and associated discussions can be found in **Methods** and **Supporting Information**.

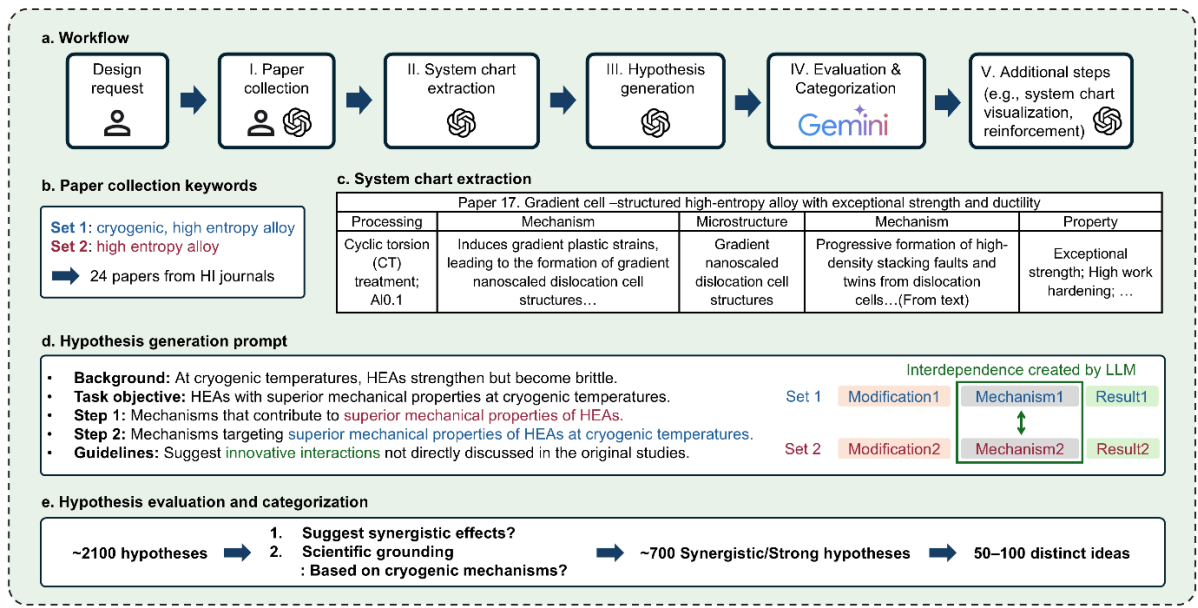

**Figure 1. Overall structure of the large-language model (LLM)-driven hypothesis generation workflow, focusing on the key concepts at each step with the example of cryogenic high-entropy alloy (HEA). a)** A schematic diagram illustrating the overall workflow. Icons indicate the entity responsible for critical information input; Human and/or LLM. **b)** Paper collection keywords. **c)** System chart extraction: Example showing one row of an extracted table summarizing a Processing-Mechanism-Microstructure-Mechanism-Property relationship (Full table: **Table S1** and prompt: **Note S2, Supporting Information**). **d)** Simplified hypothesis generation prompt; Inset: synergistic hypothesis generation through creating interdependence between mechanisms from different sources (**Note S3, Supporting Information**). **e)** Hypothesis evaluation and categorization workflow. HI: High-impact.

## 3. Results and Discussion

For cryogenic HEAs, the LLM generated ~2,100 hypotheses, classifying ~700 as synergistic and scientifically grounded, and further categorizing them into 50–100 distinct ideas (examples are in **Table 1**; the full list is in **Table S2, Supporting Information**). The ideas identified from cryogenic HEA papers predominantly involve plasticity mechanisms mediated by stacking faults (SFs), nanotwins, or the limited formation of *hexagonal close-packed* martensite. This trend aligns with recent developments in the HEA research field, which focuses primarily on single-phase HEAs or medium-entropy alloys (e.g., NiCoCr), achieving extraordinary mechanical properties at cryogenic temperatures through promoting SFs and nanotwins. Depending on specific design requirements, the keywords can be adjusted to include other alloys and a broader range of deformation mechanisms at cryogenic temperatures (e.g., using "cryogenic steel" instead of "cryogenic high entropy alloy").

Ideas from general HEA papers mainly select mechanisms involving various structural entities, such as B2 precipitates, local-chemical ordering, or texture. The hypotheses suggest using these structural entities to enhance the cryogenic plasticity mechanisms. This strategy is not yet actively explored in the field, presenting compelling directions for future research. For example, recent studies have begun leveraging coherent $L1_2$ or B2 precipitates in HEAs or steels to modulate deformation mechanisms such as SF and twin formation at cryogenic temperatures[34–36], which is similar to Ideas 6 and 11 in **Table 1**. However, it is unclear whether such references were trained in the model and influenced the generation of Ideas 6 and 11.

**Table 1.** Example ideas for designing HEAs with superior properties at cryogenic temperatures. The full list of ideas is provided in **Table S2, Supporting Information**. Individual hypotheses are provided in **Repository**-'*Cryogenic HEAs all hypotheses indexed*'.[47]

| Idea | Hypotheses | Structural entities | Core concepts |
|---|---|---|---|
| 6 | 597, 724, 791, 794, 970, … | B2-Type Ordering, Dislocations | High lattice friction from B2-type ordering hinders dislocation motion, enhancing strength. At cryogenic temperatures, this effect can promote the activation of alternative deformation mechanisms like twinning, leading to a balance between strength and ductility. |
| 8 | 366, 442, 468, 878, 905, … | Hierarchical Herringbone Microstructure, Cooperative Deformation | The hierarchical herringbone microstructure provides multiple levels for energy dissipation and crack arrest, synergizing with cooperative deformation mechanisms like twinning at cryogenic temperatures to enhance strength, ductility, and toughness. |
| 11 | 45, 411, 535, 697, 825, … | $L1_2$ Precipitates, Dislocations | $L1_2$ precipitates act as obstacles to dislocation motion, enhancing strength. This effect can synergize with other deformation mechanisms like twinning at cryogenic temperatures, leading to enhanced strength and ductility. |
| 14 | 393, 571, 596, 1269, 1286, … | Gradient Nanoscaled Dislocation Cell Structures, Deformation Mechanisms | Gradient nanoscaled dislocation cell structures, induced by various methods, influence the distribution and behavior of deformation mechanisms like twinning and stacking faults at cryogenic temperatures, enhancing strength, ductility, and toughness. |
| 21 | 133, 378, 465, 777 | SRO Domains, Stacking Faults | SRO domains impede dislocation motion and stabilize stacking faults, enhancing strength and toughness at cryogenic temperatures by preventing brittle fracture and promoting work hardening. |

Some hypotheses suggest introducing defect structures (e.g., geometrically necessary dislocations, planar dislocation glide, dislocation tangles, etc.) to facilitate cryogenic plasticity mechanisms. Notably, one hypothesis (Hypothesis 1286; Idea 14 in **Table 1**) proposes using cyclic torsion treatment to introduce gradient nanoscaled dislocation cell structures (GDSs) to accelerate the formation of SFs and twins during deformation at cryogenic temperatures. The complete hypothesis is detailed in **Figure S5a, Supporting Information**. A few key sentences taken from the hypothesis are:

- Cyclic torsion (CT) treatment induces gradient of dislocation cell structures, which serve as a foundation for the formation of stacking faults (SFs) and twins during deformation [17].
- As the temperature decreases, the low stacking fault energy (SFE) characteristic of certain high entropy alloys *further* encourages the development of *these* SFs and twins, allowing them to proliferate more easily at cryogenic temperatures [7].
- This ongoing microstructural evolution enhances both the strength and ductility of the alloy, as the material becomes progressively harder while still being able to deform plastically.

The numbers in brackets are also marked by LLM, referring to the corresponding papers cited in the hypothesis. The mechanisms combined for this hypothesis are twofold: 1) GDSs formed by CT facilitate the formation of SFs during deformation at room temperature[37] (**Figure 2a1**), and 2) at cryogenic temperatures, SFs and twins form due to reduced stacking fault energies, contributing to superior mechanical properties[38] (**Figure 2a2**). Thus, the LLM generates a new interdependency between these mechanisms, suggesting that each could influence the other's effect on SF formation for improved properties (**Figure 2b**); this synergistic effect is clearly visualized in the LLM-driven system chart in diagram format (**Figure S5b, Supporting Information**). This synergistic combination is based on two separate inferences by the model: first, that the SFs discussed in each paper represent the same structural entity, and second, that the facilitation effect of GDS on SF formation at room temperature is likely to be effective at cryogenic temperatures as well, and vice versa.

This synergistic hypothesis closely aligns with the core design idea of a recently published paper on designing HEAs with exceptional cryogenic properties,[27] which leverages GDSs produced by CT, to more actively trigger SFs at cryogenic temperatures (**Figure 2c**). This synergistic effect results in the superior strain hardenability of the HEA at cryogenic temperatures together with the increased strength. This work was published in *Science* in September 2023, after the knowledge cutoff date of the model used in this study (gpt-4-1106-preview with an April 2023 knowledge cutoff), indicating that the idea was not part of the model's training data. Thus, this example demonstrates the model's ability to bridge scientific principles from various sources to generate hypotheses that were subsequently validated in a recent publication.

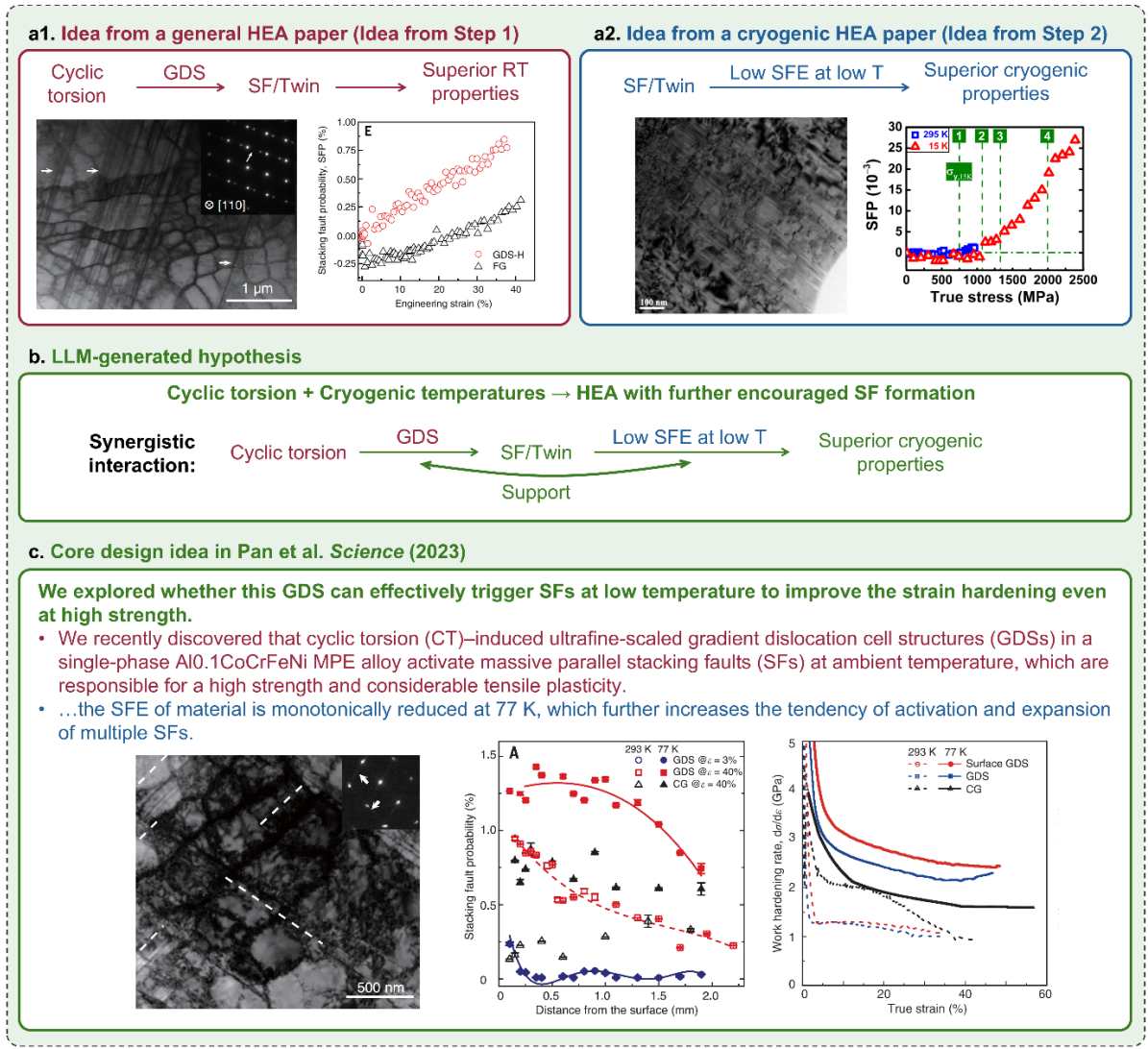

**Figure 2. Comparison of the selected LLM-generated hypothesis vs. core idea of the published paper[27] for cryogenic HEAs. a)** Ingredient mechanisms constituting the LLM-generated hypothesis. **a1)** Idea from a general HEA paper demonstrating the effect of cyclic torsion (CT) in facilitating the formation of stacking faults (SFs) and twins, leading to superior room temperature (RT) mechanical properties; GDSs: Gradient nanoscale dislocation cell structures. **a2)**

Idea from a cryogenic HEA paper demonstrating the enhanced formation of SFs at cryogenic temperatures, resulting in superior mechanical properties. SFE: Stacking-fault energy. **b)** LLM-generated hypothesis showing the synergistic interaction between the two ideas. The complete hypothesis and detailed explanations can be found in **Note S4, Supporting Information. c)** Core design idea in the published paper. Inset figures in **a1**, **a2**, and **c** present microstructure images, stacking fault-probability graphs, and the tensile curves (only in **c**) in each case, which are adapted from Refs. [37], [38], and [27], respectively. Reproduced with permission. [37] 2021, AAAS; [38] 2020, AAAS; [27] 2023, AAAS. Red represents content related to the general HEA paper, blue represents content related to the cryogenic HEA paper, and green represents the interdependency generated for the hypothesis.

Additionally, we conduct the hypothesis generation task on designing new halide solid electrolytes (SEs) with superior formability/malleability and ionic conductivity to address a challenge for all-solid-state batteries in pressure-less operation required for commercialization. A similar workflow as in the cryogenic HEAs case is used, with only minor adjustments to keywords for paper collection and prompts, keeping them general for non-specialists. Specifically, in Step III, we instruct the model to generate a "compound" rather than a "composite" structure to test the synergistic combination of mechanisms. Detailed information, including prompts and results are in **Supporting Information** (Step II: **Note S5**, **Table S3**; Step III: **Note S6**, **Table S4**). Some Ideas can be found in **Table 2**.

By requesting the generation of a compound rather than a composite, a spectrum of structural entity combinations is proposed by LLM (**Note S7, Supporting Information**). For example, some hypotheses suggest combining the benefits of a crystalline corner-sharing octahedral structure for high ionic conductivity (though typically brittle), with an amorphous structure for enhanced formability. The distribution of the two phases varies significantly among the hypotheses, ranging from multi-phase composite structures, despite instructed not to suggest composites, of the crystalline corner-sharing and amorphous phases (Hypothesis 169; Idea 59 in **Table 2**) to scientifically intricate amorphous structures, featuring crystalline corner-sharing octahedral motifs (local atomic ordering) within a globally amorphous phase, potentially achieved through controlled annealing (Hypothesis 240 in **Note S7c**, **Supporting Information)**.

**Table 2.** Example ideas for designing halide SE with superior mechanical formability/malleability with ionic conductivity in **Table S4, Supporting Information**. Idea 59 (Hypothesis 169) is in **Note S7b**. Individual hypotheses are provided in **Repository**-'*Halide SEs all hypotheses indexed*'.[47]

| Idea | Hypotheses | Structural entities | Core concepts |
|------|------------|---------------------|---------------|
| 1 | 617, 632, 654, 670, 713, … | Amorphous structure, Deep eutectic system | Combining mechanochemical milling, oxygen doping, and deep eutectic compounds creates an amorphous structure with enhanced ionic conductivity, lower activation energy, and improved formability. |
| 14 | 876, 1178 | Garnet structure, Mechanical buffer zones, Doping | A garnet structure provides inherent ionic pathways, and doping during processing creates "mechanical buffer zones" within the lattice, enhancing formability without compromising conductivity. |
| 19 | 1210 | Amorphous/crystalline halide matrix, Ball milling, Annealing | Ball milling creates an amorphous structure for enhanced ion movement and malleability, further stabilized and enhanced by annealing with dopants. |
| 42 | 1039 | Flexible crystal lattice, Polymeric/glassy matrix, Mechanochemical milling, Low-temperature processing | Mechanochemical milling of halide materials creates a disordered structure for high ionic conductivity. Low-temperature processing with a polymeric/glassy matrix enhances formability while preserving the disordered structure. |
| 59 | 169 | Orthorhombic Cmc21 space group with corner-sharing NbO2Cl4 octahedra and LiCl4 tetrahedra, Viscous/plastic matrix | Integrating a unique lattice structure with viscous/plastic features creates a solid electrolyte with high ionic conductivity and mechanical adaptability. The viscous/plastic matrix allows deformation without compromising the conductivity of the lattice structure. |

One hypothesis proposes synthesizing a *glassy halide-based electrolyte* by doping oxygen[39] and halide eutectic forming agents (e.g., $AlF_3$, $GaF_3$)[40] (**Figure S6a, Supporting Information**). Key sentences/phrases from the hypothesis include:

- This combination could lead to the ***formation of an amorphous halide-based solid electrolyte matrix***…
- mechanochemical milling with oxygen (to promote amorphization and enhance the glass network)
- The addition of oxygen…stabilizing the glass network, forming robust bridging oxygen units...
- the introduction of eutectic-forming agents according to [17]… lowering the melting point of the system and introducing more free volume and less tightly bound ions that promote fast ion transport.

- This approach would leverage the low viscosity and deformation capacity of the deep eutectic amorphous matrix, which would facilitate intimate contact with the cathode materials under minimal external pressure.

The hypothesis combines two mechanisms: 1) an oxygen-doped glassy phase (enhanced amorphization, glassy structure with improved stability from bridging oxygen units; **Figure 3a1**), and 2) a deep-eutectic amorphous phase (low melting point, enhanced formability/malleability due to low viscosity; **Figure 3a2**). This creates a novel interdependency, resulting in *a new glassy phase* that inherits characteristics from both original phases (**Figure 3b**; the visualized system chart is provided in **Figure S6b, Supporting Information**). To generate the hypothesis, several inferences are necessary: that oxygen and deep eutectic agents can collectively form a new glass phase, and that certain intrinsic properties of the original phases can be inherited in the new glass. Additionally, the model proposes that a low melting point leads to more free volume and less tightly bound ions, facilitating ion transport, which is not mentioned in both source papers. Consequently, the hypothesis suggests that oxygen doping can broaden the compositional landscape for designing novel glassy phases by promoting amorphization. This opens exciting possibilities for screening new eutectic-forming halides to discover compositions that achieve SEs with superior formability/malleability and ionic conductivity.

This hypothesis aligns with the design concept of a paper published in *Nature Energy* in September 2023[28] which was released after the end date of training for the LLM and was therefore not available to it. The paper describes doping oxygen into $LiAlCl_4$, thereby achieving a glassy inorganic SE with polymer-like viscoelasticity for pressure-less all-solid-state batteries. The paper also suggests that oxygen doping into $LiAlCl_4$ promotes bridging oxygen units (Al-O-Al), which restrict atomic rearrangement into crystals during condensation. Although not explicitly mentioned in the paper, $LiAlCl_4$ is a composition near a eutectic point in the $LiCl-AlCl_3$ phase diagram (**Figure S7, Supporting Information**), having a low melting point (~146 °C), which should be the origin of low glass transition temperatures of the glass SEs leading to viscoelasticity. Furthermore, the low glass transition temperatures are proposed as a source of high ionic conductivity through increasing free volumes.

Since the publication of the paper,[28] the development of oxyhalide-based glassy electrolytes has been drawn great attention as a promising avenue, with active research ongoing in the field.[41] Therefore, this result, along with the cryogenic HEA example, indicates that the LLM can generate novel and innovative hypotheses with the potential to impact the materials research community. Furthermore, the use of general keywords suggests that researchers not intimately familiar with the specific topics could potentially arrive at similar hypotheses.

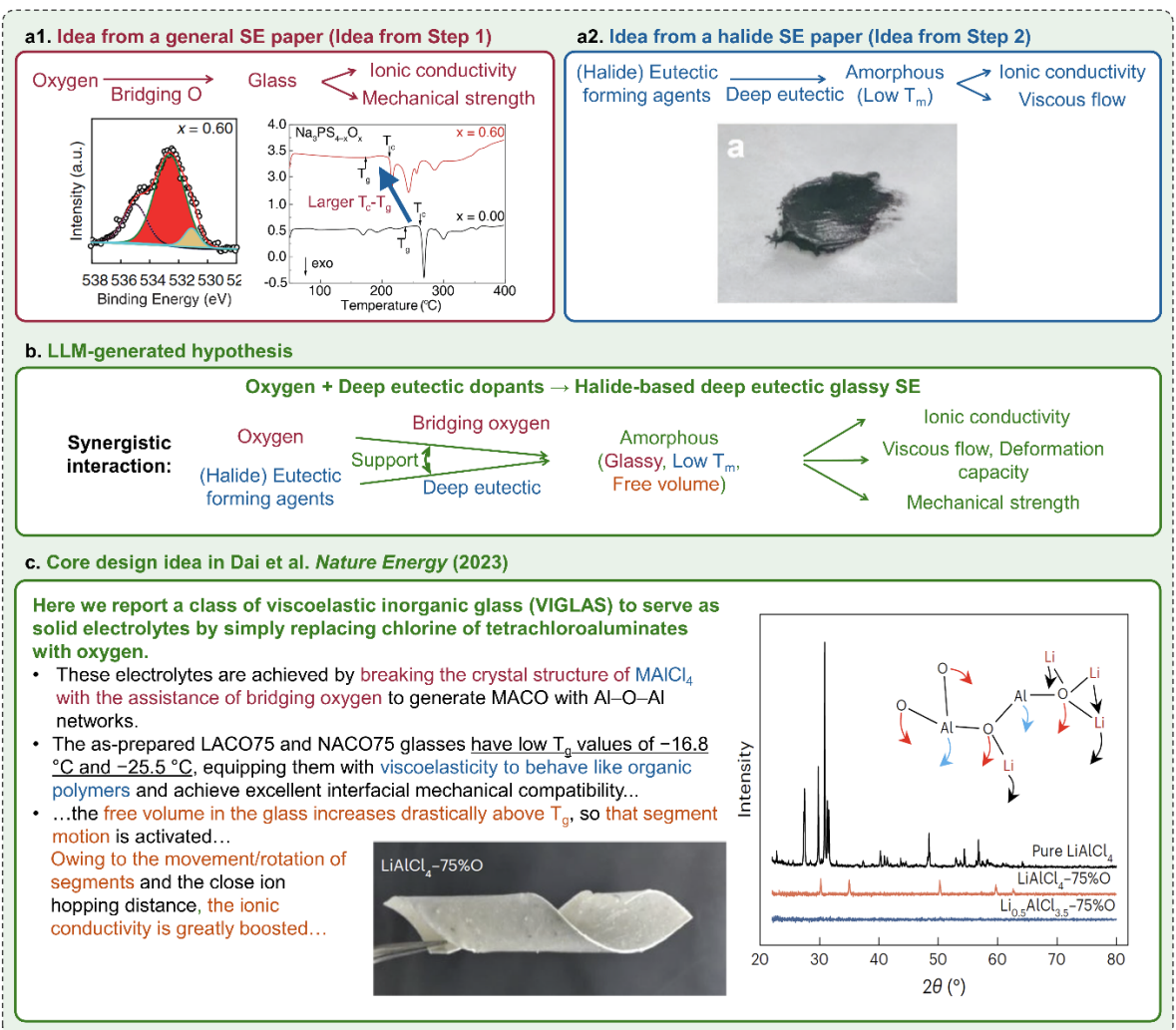

**Figure 3. Comparison of the selected LLM-generated hypothesis vs. core idea of the published paper**[28] **for halide solid electrolytes (SEs). a)** Ingredient mechanisms constituting the LLM-generated hypothesis. **a1)** Idea from a general SE paper demonstrating the effect of oxygen in stabilizing a glassy structure, leading to superior ionic conductivity and mechanical strength; Inset figures present the formation of bridging oxygen units contributing to these effects observed by O 1s X-ray photoelectron, and differential scanning calorimetry curves demonstrating oxygen's role in resisting crystallization. **a2)** Idea from a halide SE paper demonstrating the formation of an amorphous phase with a low melting point from deep eutectic dopants, resulting

in high ion conductivity and formability/malleability through viscous flow; Inset figure presents the clay-like deformed feature of the SE. **b)** LLM-generated hypothesis showing the synergistic interaction between the two ideas. **c)** Core design idea in the published paper; Inset figures present the x-ray diffraction patterns showing the formation of an amorphous structure in oxyhalide SE close to the eutectic composition, a schematic diagram illustrating bridging oxygen units, and a rolled oxyhalide SE demonstrating high formability/malleability. All inset figures are adapted with CC-BY terms for Refs. [39] for **a1** and [40] for **a2**, and reproduced with permission from Springer Nature for [28] 2023 for **c**. Red represents content related to the general SE paper, blue represents content related to the halide SE paper, and green represents the interdependency generated for the hypothesis.

**Figure 4** illustrates a recent materials design framework, the *Materials by Design* approach, now enhanced with LLM assistance.[5] Practically, materials design frameworks employ a feedback cycle from idea generation to experimental or computational validation. These frameworks are supported by advancements in experimental and computational methods, which facilitate faster and more accurate calculations, as demonstrated by density functional theory, molecular dynamics, finite element methods, and advanced synthesis and characterization techniques, thus accelerating materials design by enhancing the latter steps. However, the initial step of idea generation has traditionally been driven by researchers' intuition, posing significant challenges in pushing beyond the limits of human cognition for innovative materials design. The promising results suggest the active deployment of LLM-driven design hypothesis generation for materials design from empirical knowledge databases—one of the two pillars of materials science and engineering—even beyond the limits of researchers' knowledge. Here, we discuss some potential questions for its application in design activities:

**1. How to select keywords for literature search?** To accelerate the knowledge transfer, the literature sets can be designed to reflect multiple materials domains, processing methods, structures, or properties. This work uses two ranges of domains (cryogenic HEA vs. HEA, SE vs. halide SE). Keywords can also be generated by LLMs once the background information or objective is given.

**2. Due to the summarized nature of the system chart, is there a risk of losing key information in the intermediate materials system chart extraction step?** This does not appear

to be the case. We evaluated the accuracy of the extracted system charts from 70 papers across four sets metallurgy papers. The results, summarized in **Figure S8, Supporting Information**, show a high accuracy of over 80%, as evaluated by the authors, with the original context regarding P-S-P relationships preserved for generating hypotheses. To avoid redundancy, the details of this analysis, along with additional analyses, can be found in **Methods** and **Note S8 and S9**, respectively.

**3. Can LLMs fairly evaluate the synergistic and scientifically grounding of hypotheses?** We manually evaluated synergy and scientific grounding of the 200 hypotheses and used the results to refine the model's evaluation prompts, achieving ~80% accuracy (See **Methods** for details). It is important to note that the prompts and model (gemini-1.5-pro) used in this work were primarily employed to demonstrate the validity of the proposed approach. We believe that with the next generation of LLMs, prompts, and potentially fine-tuning, significantly greater advancements can be achieved.

**4. What is the next step of hypothesis generation by LLM?** As seen in **Figure 4**, hypotheses can be reinforced in the next step by integrating external computational tools or LLM-based question-answering using retrieved related literature for detailed planning. Specifically, since the hypotheses involve a few interplays between P-S-P, applying them to design projects requires comprehensive plans that incorporate the entire P-S-P relationships of the potential materials. An additional process can be utilized to incorporate the ingredient papers, as there may be less essential P-S-P information that, while not extracted for building system charts in Step II, could still be useful for improving the comprehensiveness of the hypotheses-based system charts. See **Note S4 and S6, Supporting Information** for examples of detailed explanations generated by the LLM in this step for the present cryogenic HEA and halide SE hypotheses, which even include additional novel hypotheses observed and highlighted in the aforementioned recent papers.

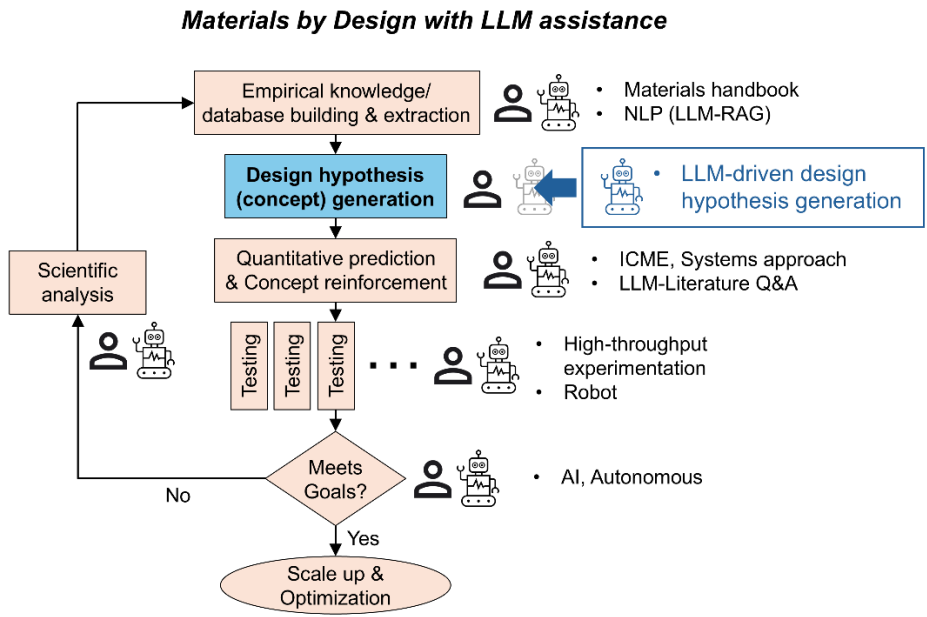

**Figure 4. Materials by Design approach with LLM assistance.** Modified from McDowell et al[5] to incorporate LLM. Our proposal is highlighted in blue. The materials handbook offers fundamental knowledge and parameters for building design concepts. The Integrated Computational Materials Engineering (ICME) approach significantly reduces the number of trials by providing initial validation of concepts through various simulation methods, including uncertainty prediction. The development of high-throughput testing methods has greatly increased the amount experimental data available. More recently, autonomous design concepts, which rely on artificial intelligence to determine the next testing parameters, have further reduced the number of experiments needed. LLMs, in addition to guiding external tools and scientific analysis, can generate design hypotheses that have the potential to direct the entire design process. NLP: Natural Language Processing; RAG: Retrieval Augmented generation; Q&A: Question and Answering.

## 4. Conclusion

In summary, this study shows that LLMs can be used to generate novel scientific hypotheses for materials design on par with those that underlie papers in prestigious journals in the field. This highlights the growing potential of LLMs to provide valuable scientific insights, even as the ability of LLMs to fully reflect the complexities of the physical world or perform common-sense reasoning remains a topic of active research.[42–44] Historically, materials science has occupied a unique position by dealing with materials of complexity that lies between the realms of individual atoms and complex organisms. [1] It has advanced through a combination of both scientific and empirical information,[4] often successfully represented through linguistic representations. [45,46] Thus, we believe that materials science is an ideal domain to benefit from LLM advancements. We envision the continued evolution of this approach paving the way for a new era of materials design, where LLMs accelerate and enrich discovery and democratize design activities.

## 5. Methods

*Model related information:*

Both OpenAI ChatGPT API and Gemini API are used within Python 3.10.8, with the 'openai' package version 0.28.1 and the 'google-generativeai' package version 0.7.2, respectively. For all work other than hypotheses evaluation, gpt-4-1106-preview snapshot model of GPT-4 is utilized, while gemini-1.5-pro is used specifically for hypotheses evaluation.

For gpt-4-1106-preview, temperature = 0.0 is maintained for all tasks except for hypotheses generation, where temperature = 1.0 is applied. When the temperature is set to 0.0, the model's responses are more deterministic and focused, leading to predictable and reliable outputs. Conversely, a temperature of 1.0 allows for more creative and varied responses, which are particularly useful in generating diverse hypotheses. The maximum output token limit is set to be 4000. The system messages are tailored according to the field of study. For tasks related to alloys, the system message is set to "You are an expert in the alloy field of Materials Science and Engineering". For tasks associated with lithium batteries, the system message is set to "You possess expertise in the field of all-solid-state Lithium battery research". All other parameters are kept at their default settings, which were: frequency_penalty = 0.0, presence_penalty = 0.0, top_p = 1.0, logprobs = *False*, stream = *False*, n = 1, logit_bias = *null*, stop = *null*.

For gemini-1.5-pro, temperature = 0.0 and max_output_tokens = 4000, are maintained for all tasks. All other parameters are kept at their default settings, which were: top_p = 0.95, top_k = 64.

*System chart extraction:*

Materials system charts are extracted from individual papers. The workflow is shown in **Figure S2, Supporting Information**, and the prompt is provided in **Note S2 and S5, Supporting**

**Information**. The model first extracts the targeted properties, followed by the (micro)structures relevant to these properties; the terms 'structure' and 'microstructure' are used interchangeably in this article. Next, the mechanisms connecting the structures and target properties are extracted. The model can reference mechanisms not mentioned in the paper, if they are indicated as "From knowledge base". This extracted Property-Mechanism-Structure information then forms a sub-table (Sub-table 1 in **Figure S2a, Supporting Information**). Next, the model extracts the processing methods important for achieving the predetermined structures and their associated mechanisms (Sub-table 2). As both processing and properties are tied to structure, the two sub-tables are then connected to form the entire P-M-S-M-P of a system chart; the number of rows in the chart is equal to the number of extracted structures. Through this approach, the information mutually crucial for both scientific cause-and-effect (from processing-structure-properties) and engineering goals-means (from properties-structure-processing) perspectives[1,2,4] can be effectively extracted.

*Prompt engineering for system chart extraction:*

During the prompt engineering process, we continuously assess the quality of the generated system. We select 70 papers across four datasets to optimize and evaluate the system chart extraction process (**Repository**[47]). First, the initial prompt is built using ten experimental papers from high-impact journals for their well-organized information. Second, ten computational papers are included to extend the prompt's capabilities to simulation-rich studies. These two sets iteratively refine prompts until the average HMI and average mechanism score exceed 0.8. Third, 30 papers on cryogenic HEAs are chosen to test extraction related to mechanical behavior. Lastly, 20 papers on tungsten alloys for fusion applications are selected to test the chart extraction across various environments and target properties, including oxidation, transmutation, and different processing routes like sintering.

The system charts in table format are evaluated using two metrics: the human machine-readability index (HMI)[48] and a mechanism score specifically defined for this study (**Figure S7a, Supporting Information**). HMI involves manual evaluation of the accuracy and relevance of content in the P-S-P columns. The mechanism score assesses the mechanistic accuracy of the extracted mechanisms and source labeling accuracy in the Mechanism columns. Mechanism fidelity counts mechanisms with both accurate mechanistic explanation and correct labeling (PP in the figure).

For HMI, the criteria include 1) Incorrect actions: Fabricated definitions, inaccuracies, irrelevant information; 2) Partially correct actions: Accurate content miscategorized, mixed correct and incorrect elements; 3) Correct actions: Precise information mentioned in the articles. After counting the number of occurrences of each action, HMI can be calculated as follows:

$$HMI\ (\%) = \frac{I(0)+PC(0.5)+C(1)}{A_T} \times 100,$$

where $I$ = incorrect actions $PC$ = partially correct actions $C$ = correct actions $A_T$ = total actions. Moreover, if the table does not include the core ideas of the article (although a subjective criterion) 20% is deducted from the HMI.

For the mechanism score, the criteria include: 1) Labeling accuracy: Whether the mechanism is correctly classified as "From text" or "From knowledge base" based on its source; 2) Mechanistic accuracy: Whether the mechanism is scientifically reasonable and correctly illustrates the relationship between components. For "From text" mechanisms, accuracy is evaluated based on statements in the original research article, and for "From knowledge base" mechanisms, accuracy is evaluated based on the authors' knowledge; 3) Mechanism fidelity: The number of mechanisms that are both mechanistically sound and labeling accurate.

Results, summarized in **Figure S7b, Supporting Information**, show that all scores for the four datasets exceed 0.8; Detailed evaluation examples and complete results are provided in **Table S5, Supporting Information** and **Repository**,[47] respectively. This validation process demonstrates the effectiveness and accuracy of our present prompt engineering approach, with high similarity between machine-generated outputs and expert evaluations. We anticipate further improvements using advanced LLM techniques, involving causal analysis,[49,50] fine tuning,[51,52] and ontology integration,[53,54] which will be explored in future work.

*System chart visualization:*

To convert system charts from individual hypotheses or papers in table format into diagrams, a series of prompts are used (**Note S10 and S11, Supporting Information**). Initially, all microstructures in the table are tagged based on their categorization results. Detailed parameters are then simplified, while retaining environmental conditions. Processing methods and composition/key elements, along with their corresponding mechanisms, are separated into individual rows if they were initially combined. Multiple properties in one row are also separated into different rows. Finally, 'N/A' values in the Structure and Property columns are identified and replaced by comparing relevant values with other rows. Generating diagrams from system charts is done within Python 3.10.8. The libraries `matplotlib`, `networkx`, and `textwrap` are utilized for plotting, network creation, and text wrapping for better visualization.

*Hypothesis evaluation:*

To assess the presence of synergistic effects in each hypothesis, we establish a universal prompt (**Figure S4a, Supporting Information**; the complete prompt is provided in **Note S12, Supporting Information**), classifying hypotheses as 'Synergistic' or 'Additive.' A 'Synergistic' hypothesis is identified in both cases of cryogenic HEA and halide SE when there is generated interdependence between the two combined mechanisms or when the combined action of the two mechanisms influences or creates a specific structure. In contrast, an 'Additive' hypothesis lacks any direct interaction or interdependence between mechanisms. Based on these criteria, each hypothesis is scored on a 1–5 scale. Hypotheses scoring above 3 are classified as 'Synergistic', while those scoring 3 or below are classified as 'Additive'. For 'Synergistic' hypotheses, sentences illustrating the interdependence between mechanisms and core structures connecting synergistic mechanisms are highlighted for the subsequent hypothesis categorization process.

To assess the level of scientific grounding for each hypothesis, we examine the alignment between our stated materials design goals and the actual content of the hypothesis. We employ general prompts (**Figure S4b, Supporting Information**; The complete prompt is provided in **Note S13, Supporting Information**) tailored for cryogenic HEA and halide SE. A 'Strong' hypothesis in the case of cryogenic HEA demonstrates strong scientific grounding by explicitly harnessing phenomena or mechanisms specific to cryogenic conditions, whereas a 'Weak' hypothesis lacks this specificity. In the case of halide SE, a 'Strong' hypothesis exhibits strong scientific grounding by incorporating mechanisms that contribute to formability/malleability, a feature absent in a 'Weak' hypothesis.

Some examples are provided below (full hypotheses of these examples are in **Note S14, Supporting Information**.):

Weak/Additive (Hypothesis {189}): *Coherent nanoprecipitates effectively block dislocations and impart high yield strength, while deformation-induced stacking faults allow for accommodating plastic deformation.*

Strong/Synergistic (Hypothesis {29}): *At cryogenic temperatures, the mechanical properties of HEAs are significantly altered due to the increased activity of twinning and the formation of stacking faults… At cryogenic temperatures, the resistance to dislocation movement due to precipitates can induce a higher stress threshold for dislocation glide, which favors twinning as an alternative deformation mechanism.*

To test the effectiveness of the method, the authors also manually evaluated 200 cryogenic HEA hypotheses and compared the results with the model's evaluation. The manual evaluation results and the comparison results are provided in **Repository**[47] and **Table S6, Supporting Information**, respectively. The model exhibits satisfactory accuracy (0.79), precision (0.70), recall (0.83), and F1 (0.76) for synergy and accuracy (0.83), precision (0.96), recall (0.82), and F1 (0.88) for scientific grounding. Compared to scientific grounding, the precision for synergy is lower. This is largely due to the model's tendency to over-interpret the presence of synergistic effects that are not explicitly stated in the hypothesis, often assuming their existence even when not clearly indicated.

*Hypothesis categorization:*

To obtain unique ideas from a large number of hypotheses and reduce the difficulty for human researchers in finding their desired hypotheses, an inclusive hypothesis categorization prompt (**Note S15, Supporting Information**) is utilized after evaluation. For hypotheses evaluated as both 'Strong' and 'Synergistic,' they are grouped into five equal-length chunks, and initial categorization is performed on each chunk. Factors considered in this step include the combined paper numbers (referencing the specific papers used to derive the hypothesis), core structural entities, and synergistic mechanism sentences identified through synergy evaluation. Hypotheses exhibiting similarities in these elements are likely to share content and are merged into a single idea.

Following the generation of ideas for each of the five chunks, final categorization is applied to the combined results. This second step also focuses on merging ideas with similar combination numbers, core structural entities, and concepts. To avoid an overwhelming number of generated ideas, the categorization process is automatically halted by the code once the number of existing ideas exceeds 50 (this check occurs immediately after each LLM response). While the initial categorization targets individual hypotheses, the final categorization operates on the ideas generated in the previous step.

We observe a significant loss of hypotheses during the second step; often, 1/3 to 1/2 of the hypotheses are overlooked in this process. However, this two-step categorization process is necessary due to current constraints on LLM output length; obtaining all final categorization results in a single response is not always feasible. An additional prompt (**Note S15, Supporting Information**) is employed to maintain the conversation flow. We anticipate that advancements in LLMs will enable the effective categorization of hypotheses in a single step.

**Supporting Information**

Supporting Information is available from the authors upon request.

**Acknowledgements**

This work was supported by the start-up funds from the University of Wisconsin-Madison provided by the Wisconsin Alumni Research Foundation (WARF). D.M. and M.P.P. acknowledge the support from National Science Foundation Cyberinfrastructure for Sustained Scientific Innovation (CSSI) Award No. 1931298.